\documentclass[letterpaper, 10 pt, conference]{ieeeconf}  

\IEEEoverridecommandlockouts                              

\usepackage{blindtext}

\usepackage{graphics} 
\usepackage{booktabs}
\usepackage{cite}
\usepackage{balance}

\usepackage{graphicx} 
\usepackage{verbatim}
\usepackage{xcolor}
\usepackage{subcaption}

\usepackage{hyperref}

\usepackage{color}
\usepackage{colortbl}
\definecolor{todo-red}{RGB}{200,12,12}
\definecolor{green4}{RGB}{0,128,0}

\newcommand{\reffig}[1]{Fig.~\ref{#1}}

\newcommand{\reftab}[1]{Table~\ref{#1}}
\newcommand{\refsec}[1]{Section~\ref{#1}}

\usepackage{xcolor}

\title{\LARGE \bf
Topomap: Topological Mapping and Navigation \\ Based on Visual SLAM Maps
}

\author{Fabian Bl\"ochliger, Marius Fehr, Marcin Dymczyk, Thomas Schneider and Roland Siegwart
\thanks{All authors are with the Autonomous Systems Lab, ETH Zurich. Contact: {\tt\small fabian.bloechliger@mavt.ethz.ch}.}
}

\begin{document}

\maketitle
\thispagestyle{empty}
\pagestyle{empty}

\begin{abstract}

Visual robot navigation within large-scale, semi-structured environments deals with various challenges such as computation intensive path planning algorithms or insufficient knowledge about traversable spaces.
Moreover, many state-of-the-art navigation approaches only operate locally instead of gaining a more conceptual understanding of the planning objective.
This limits the complexity of tasks a robot can accomplish and makes it harder to deal with uncertainties that are present in the context of real-time robotics applications.

In this work, we present \emph{Topomap}, a framework which simplifies the navigation task by providing a map to the robot which is tailored for path planning use.
This novel approach transforms a sparse feature-based map from a visual Simultaneous Localization And Mapping (SLAM) system into a three-dimensional topological map.
This is done in two steps.
First, we extract occupancy information directly from the noisy sparse point cloud.
Then, we create a set of convex free-space clusters, which are the vertices of the topological map.
We show that this representation improves the efficiency of global planning, and we provide a complete derivation of our algorithm.
Planning experiments on real world datasets demonstrate that we achieve similar performance as RRT* with significantly lower computation times and storage requirements.
Finally, we test our algorithm on a mobile robotic platform to prove its advantages.
\end{abstract}

\section{Introduction}

\begin{figure}
    \centering
    \begin{subfigure}[b]{0.99\columnwidth}
        \centering
        \includegraphics[width=1.0\columnwidth]{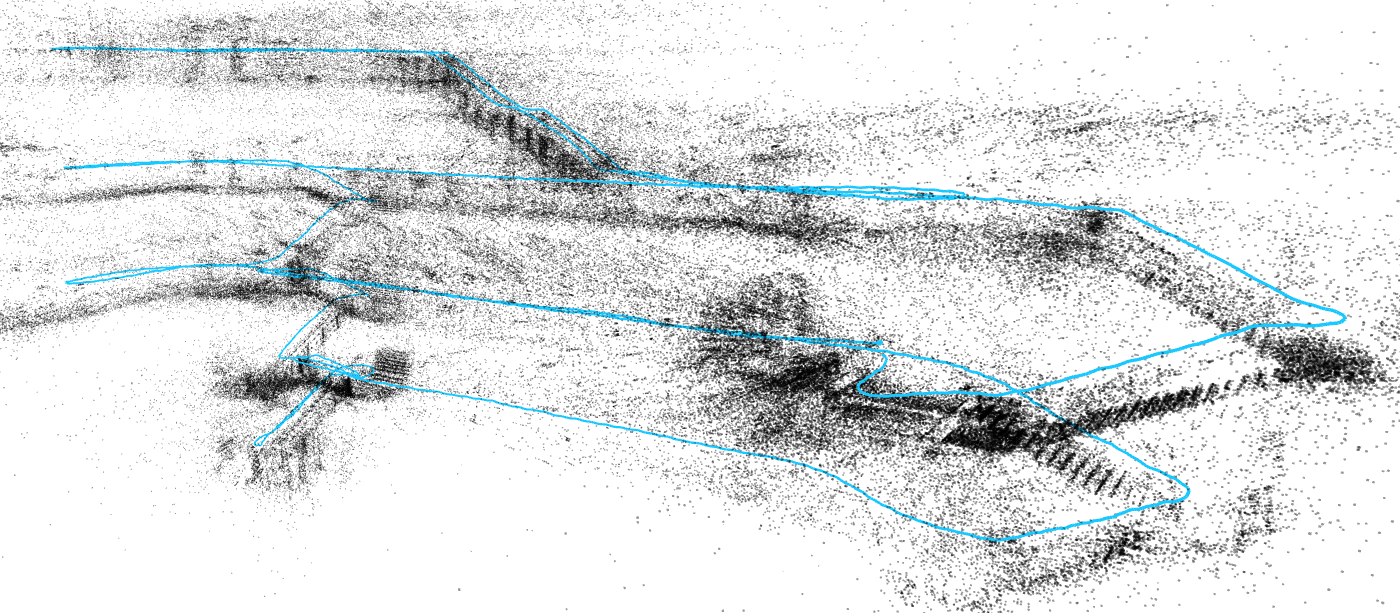}
        \caption{Visual SLAM map.}
        \label{fig:teaser_image:slam_map}
    \end{subfigure}
    
    \begin{subfigure}[b]{0.99\columnwidth}
        \centering
        \includegraphics[width=1.0\columnwidth]{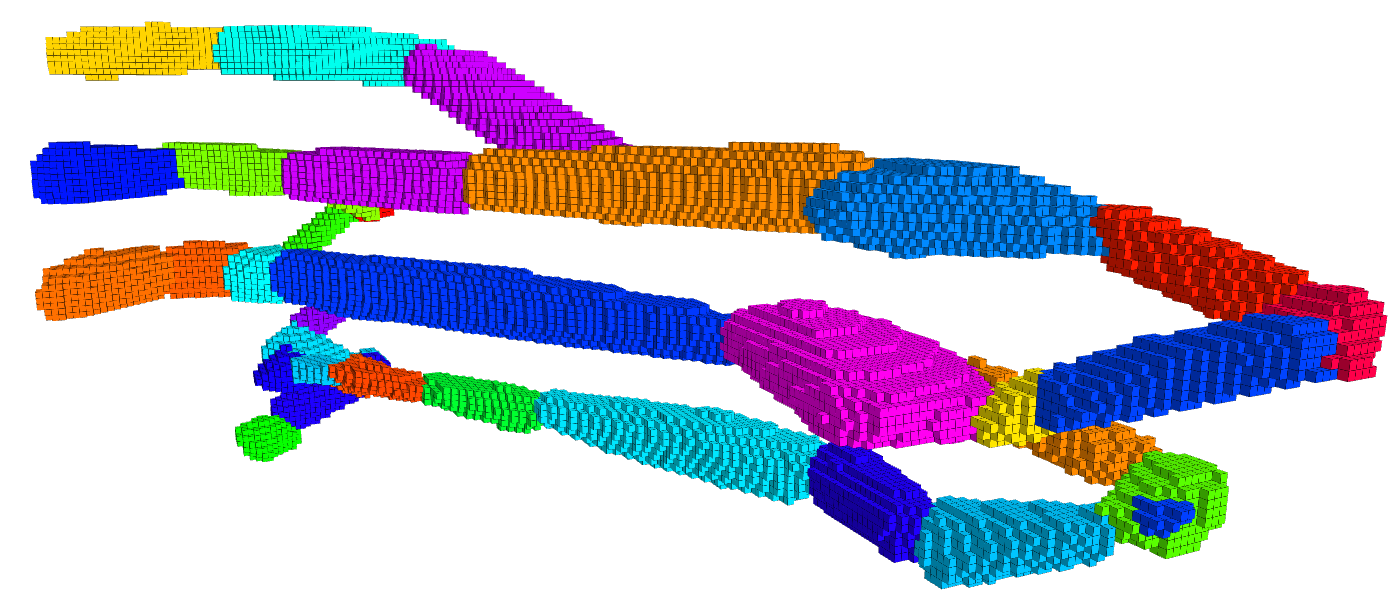}
        \caption{Topological map (convex voxel clusters).}
        \label{fig:teaser_image:topo_map}
    \end{subfigure}
    
    \begin{subfigure}[b]{0.99\columnwidth}
        \centering
        \includegraphics[width=1.0\columnwidth]{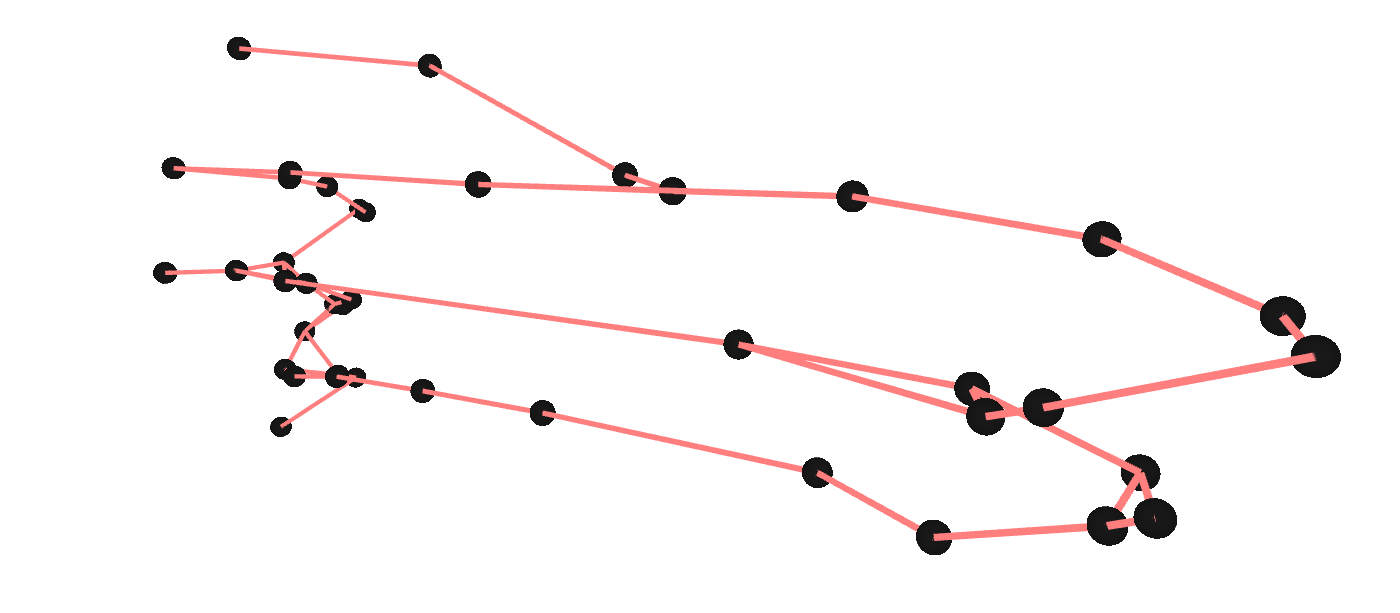}
        \caption{Navigation graph.}
        \label{fig:teaser_image:navigation_graph}
    \end{subfigure}
    \caption{Our proposed topological mapping approach applied to a multi-floor indoor scene:
    (a)~The input to our framework is a sparse visual SLAM map of the environment with $423'000$ triangulated 3D landmarks.
    The landmark positions are estimated based on the multi-view geometry and the feature tracking pipeline of a visual-inertial odometry estimator.
    The length of the trajectory is $360\,\mathrm{m}$.
    (b)~Using free space information extracted from the landmarks, we build a topological map consisting of convex voxel clusters (vertices) and their adjacent areas (edges), which we call \emph{portals}.
    The vertices denote free, traversable areas within the environment.
    (c)~The derived navigation graph makes global path planning within the explored environment easy and computationally inexpensive.
    }
    \label{fig:teaser_image}
\end{figure}

Mobile robots have recently left the research laboratories and are becoming more and more ubiquitous.
In many of the resulting applications, including those aimed at the consumer market, navigation is a key capability.
It is hard to imagine robotic vacuum cleaners or surveillance robots without at least a minimal suite of navigation and path planning skills.
Another rapidly developing market are the Augmented Reality (AR) and Virtual Reality (VR) applications, where it is often expected that a mobile device will be able to guide a user to a certain location.
In both of these use cases, reliable navigation within a global coordinate frame is crucial and ideally requires a minimal sensor setup and limited computational resources.

State-of-the-art navigation and path planning approaches are often based on an occupancy map representation~\cite{Elfes1989}.
Then, a planning algorithm, such as RRT~\cite{Lavalle98} or variants thereof, can be deployed to obtain a global path within the free space of the given map.
Occupancy maps, however, are usually built using either expensive laser sensors~\cite{Hornung2013}, RGB-D cameras~\cite{Endres20143} or computationally demanding stereo cameras~\cite{Burri2015}.
Additionally, an occupancy map representation does not provide a higher-level understanding of the environment, for example division of the free space into separable parts (e.g. rooms in a building).
Finally, planning and navigation using occupancy maps is a computationally demanding problem and can be challenging in the presence of noisy input data.
This limits the capabilities of many mobile platforms.

This work introduces \emph{Topomap}, a lightweight topological mapping and navigation approach which addresses the aforementioned disadvantages.
It represents a complete and simple solution for autonomous navigation within large-scale visual SLAM maps, where topological maps are derived directly from triangulated 3D positions of the visual landmarks.
This implicitly leads to a tight coupling of the localization and navigation map.
Our approach solely relies on a standard monocular camera that is lightweight, small and can easily be placed on most robotic platforms.
The working principle of our framework is demonstrated in \reffig{fig:teaser_image}.
We divide the environment into a set of free space clusters which correspond to the vertices of the topological map. 
We enforce the convexity of their shape and as a result, a robot can cross each of those regions without the risk of running into a static obstacle.
On the other hand, the portals (adjacency regions) are the places where a safe transition from one to another vertex is possible.

\begin{figure}
    \centering
    \begin{subfigure}[b]{0.9\columnwidth}
        \centering
        \includegraphics[width=1.0\columnwidth]{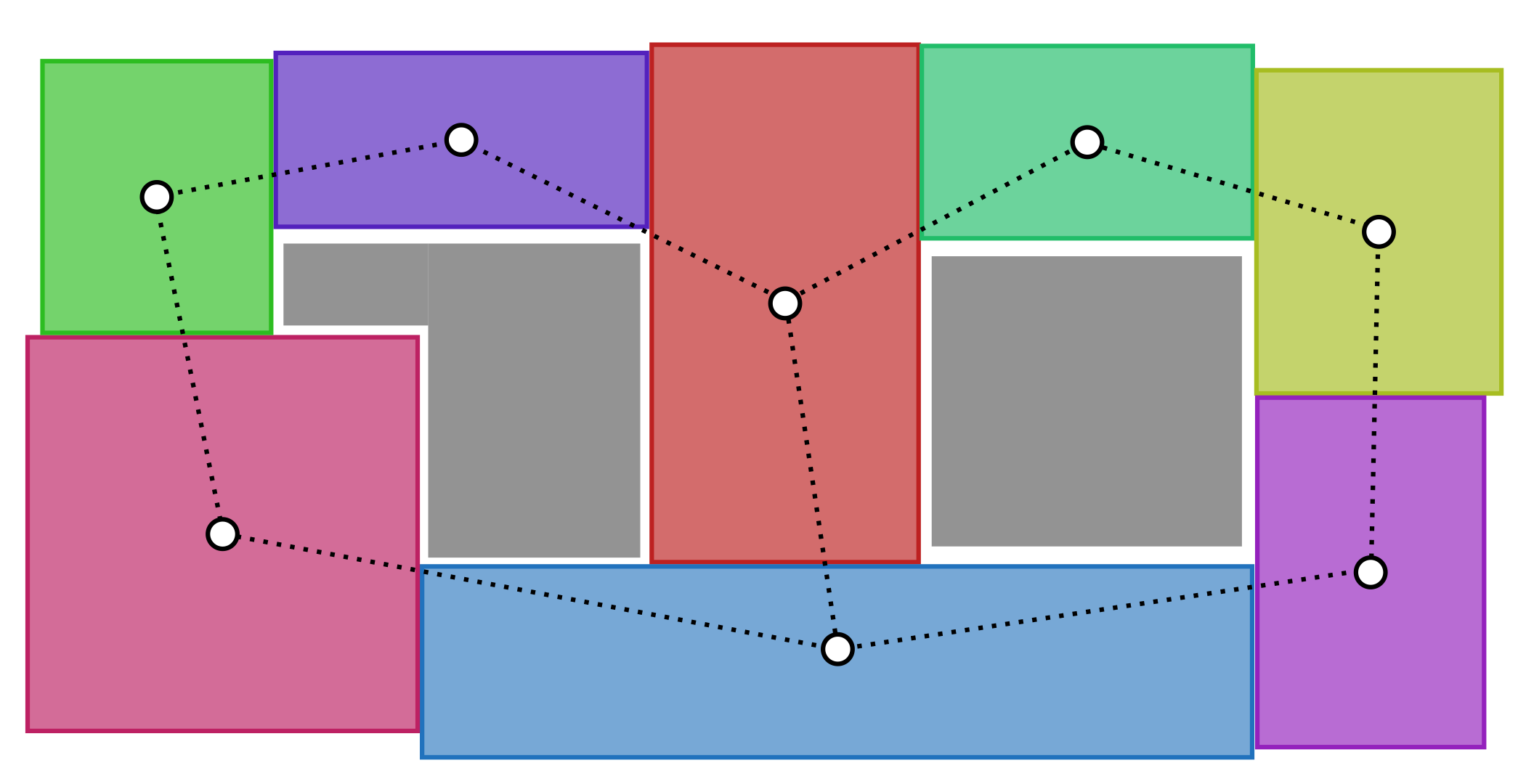}
        \caption{Topological graph (connectivity of clusters).}
        \label{fig:example_idea:topo_graph}
    \end{subfigure}
    
    \begin{subfigure}[b]{0.9\columnwidth}
        \centering
        \includegraphics[width=1.0\columnwidth]{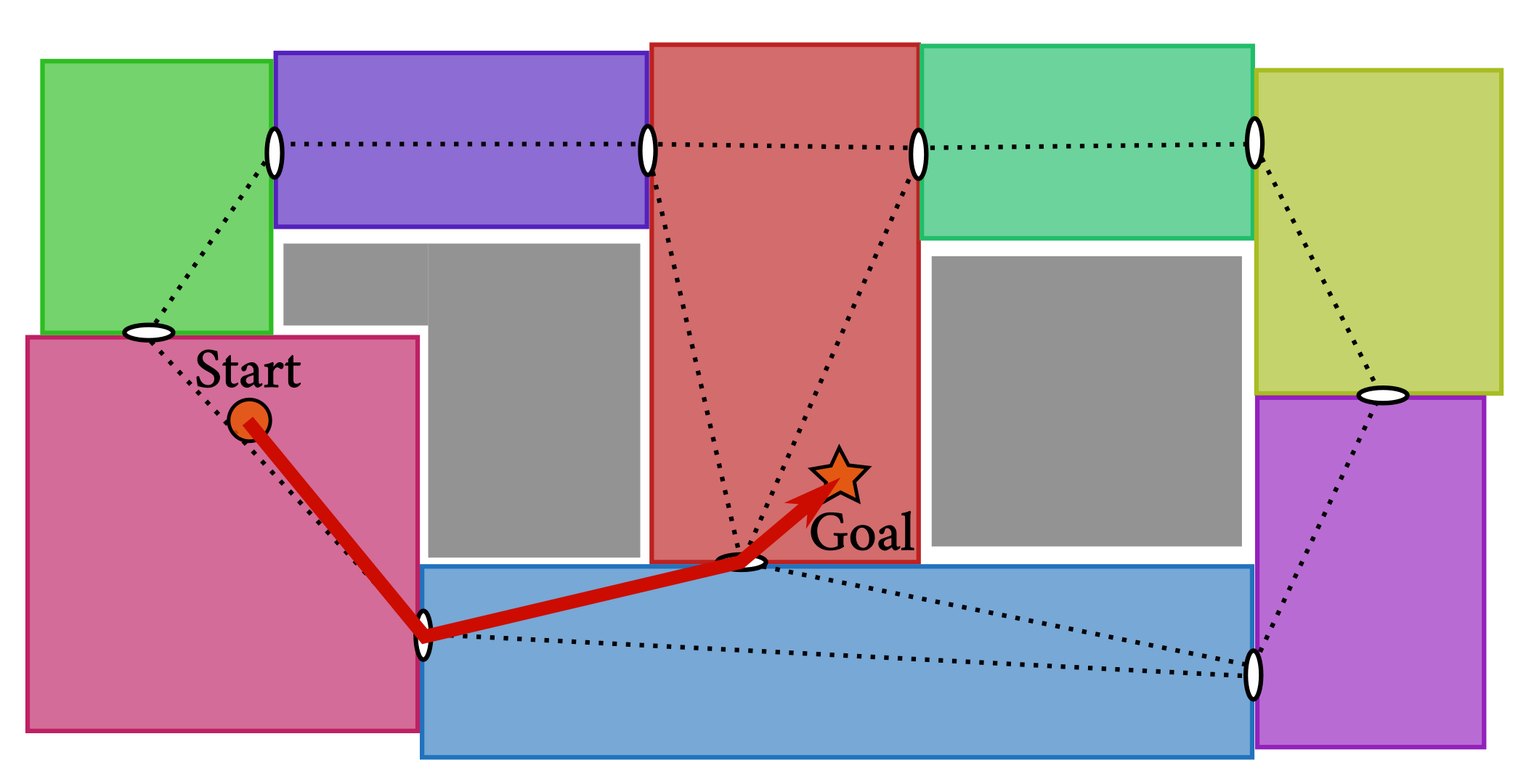}
        \caption{Navigation graph (connectivity of portals).}
        \label{fig:example_idea:navi_graph}
    \end{subfigure}

    \caption{The two basic elements of the topological map:
    (a)~The \emph{topological graph} contains the relation between the convex free space clusters (vertices):
    All adjacent vertices are connected by a topological edge, which indicates that a robot can directly move between the corresponding vertices.
    (b)~The \emph{navigation graph} is the dual graph of the topological graph. It is obtained by connecting all portals of each topological vertex.
    In order to keep the navigation approach simple, we only use the centers of the portals in our experiments.
    An example of an A-to-B path planning task is shown.
    }
    \label{fig:example_idea}
\end{figure}

Having divided our free space into a set of convex clusters, we can derive two graphs which are shown in \reffig{fig:example_idea}. The topological graph (\reffig{fig:example_idea:topo_graph}) holds the connectivity of the convex voxel clusters (vertices), whereas its dual graph, the navigation graph (\reffig{fig:example_idea:navi_graph}), can be used for path planning by any graph based planning algorithm.
To the best of our knowledge, \emph{Topomap} is the first system which is designed to extract free space from sparse visual features in order to create a topological map representation of the environment.
By using sparse features, efficient 3D structures and a simple navigation concept, our algorithm can be deployed on mobile platforms with limited computational resources.

The contributions of this work can be summarized as follows:
\begin{itemize}
    \item We propose to use a sparse visual SLAM map to create a reliable free-space representation and a subsequent topological map of the area. 
    \item We present an entire processing pipeline that takes a visual map as input and creates a topological map that can be used by a global planner.
    \item We propose an algorithm that employs -- based on a volumetric occupancy grid -- voxel cluster growing and merging to generate convex free space clusters from noisy and partly incomplete visual SLAM data.
    \item We present an extensive evaluation of the framework using real life datasets with different topological characteristics and compare our navigation concept to a state-of-the-art grid based planner.
\end{itemize}

\section{Related work}

Topological mapping combined with vision sensors has a long tradition in mobile robotics \cite{Kortenkamp1994}.
In the context of SLAM, there are multiple reasons to partition an environment into a number of discrete places:
Past works include topometric localization~\cite{Badino2011}, the use of a hierarchical bundle adjustment~\cite{Lim2012} or map reduction purposes~\cite{Dymczyk2015}.

One idea related to our proposed algorithm is to construct topological maps on top of 2D grid-based maps by dividing the free space into disjoint regions.
The regions are delimited by narrow passages derived from the environment's Voronoi decomposition~\cite{Thrun1998}, which can also be done in an incremental fashion~\cite{Liu2014}.
%
However, we believe that applying Voronoi diagrams in the context of visual SLAM maps is challenging, as we do not have accurate 2D/3D laser maps, but noisy and sparse 3D landmarks.

Another widely spread idea is to divide a map into meaningful keyframe or landmark clusters based on a similarity measure, e.g. landmark co-visibility~\cite{Zivkovic2005, Blanco2006, Fraundorfer2007, VazquezMartin2009}.
This approach might result in meaningful clusters from a human point of view (e.g., entering a new room results in a new group of keyframes), but co-visibility clusters have no concept of free space or convexity, which we believe to be key components for efficient path planning.
Enforcing convex free space clusters guarantees that the robot can move freely within each cluster.
Furthermore, a typically low viewpoint invariance of visual features will tend to create two regions of a single place which has been traversed from different directions when using co-visibility based methods.

The third group of approaches to topological mapping is attaching local occupancy grids at different places along the metric SLAM map~\cite{Konolige2011}.
The key part of those algorithms is the strategy to select places to store the grid. 
This can be done, for example, by using fixed size, overlapping cubes~\cite{Schmuck2016}.
These approaches may help to capture the topology of an environment.
However, they only partially simplify the navigation process as a local path needs to be planned through the topological vertices, whereas we have a strong prior assumption about the convex shape of the vertices.

The works which are closest to ours but more targeted towards simplifying polynomial trajectory generation for MAVs are \cite{Deits2015Planning} and \cite{Liu2017}, which generate large overlapping convex regions~\cite{Deits2015Iris} and compute a path through this regions which can be followed by a MAV.
Instead of using many overlapping clusters, we propose to use a compact expansion step in the cluster creation to capture more of the local free space and reduce the total number of clusters.
Furthermore, our voxel based cluster creation algorithm allows a seamless integration with discrete occupancy maps from real world data.

\section{Methodology}

This section introduces the methodology of the proposed algorithm.
In \refsec{sec:map_representation}, the topological map representation is explained.
Then, in \refsec{sec:occupancy} we describe in detail how we can derive free space information solely from sparse visual SLAM features.
In the next step, presented in \refsec{sec:cluster_growing}, we grow compact, convex free space clusters using the information computed beforehand.
Neighboring clusters are then merged if their combined shape includes a low number of obstacle voxels, as described in \refsec{sec:cluster_merging}.
This reduces the number of clusters and results in a topological map whose vertices are convex free space regions.
In \refsec{sec:topo_navigation}, we show how to use the obtained topological map for global path planning.

\subsection{Topological Map Representation}
\label{sec:map_representation}

We propose to construct a topological map by segmenting the free space of the entire environment into a a set of convex regions (topological vertices).
As a result, each vertex corresponds to a certain partially enclosed area within the environment (e.g. a room) which is connected to neighboring vertices by portals.
This topological map representation resembles the way humans perceive the environment when navigating \cite{lynch1960image}, and is likewise convenient from a robot's planning perspective.

The convex regions are represented as clusters of voxels, which means that they can be directly derived from voxel based occupancy maps.
The occupancy maps, however, do not need to be stored, which significantly reduces the storage requirements compared to state-of-the-art approaches.
Instead, we only serialize the convex hull of the regions corresponding to the topological vertices.
This way, the useful information of our topological map is preserved (vertex volume, topological connections and portals).
The serialized data is sufficient to deduce in which region a given position is located.

\subsection{Occupancy from Sparse Features}
\label{sec:occupancy}

\emph{Topomap} extracts discrete approximate free space information from sparse landmarks by using voxel based Truncated Signed Distance Fields~\cite{Newcombe2011}.
TSDFs are commonly used in combination with depth sensors (e.g. laser rangefinders or densified multi-camera setups) \cite{Oleynikova2016}.
We will, however, focus on using sparse visual SLAM features for this purpose.

Our first steps will be analogous to fusing depth measurements into a volumetric TSDF grid using the traditional sensor modalities, i.e. by ray tracing the 3D grid from sensor origin to the measured 3D point.
Hence each triangulated 3D landmark present in the SLAM map is ray traced from its observer pose.
The distance values in all traversed voxels are updated according to the distance to the landmark up to a pre-defined maximum distance value (truncation distance).
Ray tracing observations of all 3D landmarks will result in a voxel map which contains projective distances to obstacle surfaces, which is in fact only an approximation to the real distance. 
The TSDF construction step is provided by the volumetric mapping library \emph{voxblox}~\cite{Oleynikova2017}.

The TSDF representation based on noisy and sparse visual features requires some additional post-processing to obtain reliable information about the voxel occupancy.
First of all, we binarize the information by thresholding the distance value of each voxel (we chose $90\%$ of the truncation distance).
Secondly, we propose a subsequent filtering step which removes small occupied voxel groups which are not connected to any other occupied part. 
These outliers might come from dynamic objects while building the SLAM map or badly triangulated landmarks.

\subsection{Compact Cluster Growing}
\label{sec:cluster_growing}

\begin{figure}[tpb]
    \begin{center}
        \begin{minipage}[t]{0.35\columnwidth}
            \centering
            \includegraphics[width = \textwidth]{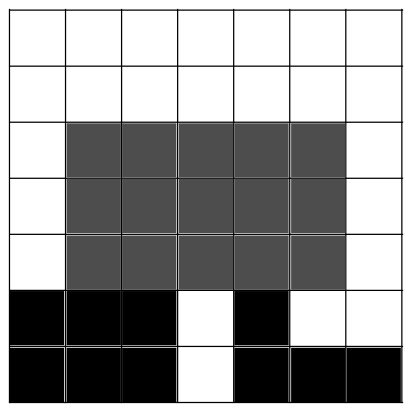} \\ (a)
        \end{minipage}
        \hfill
        \begin{minipage}[t]{0.35\columnwidth}
            \centering
            \includegraphics[width = \textwidth]{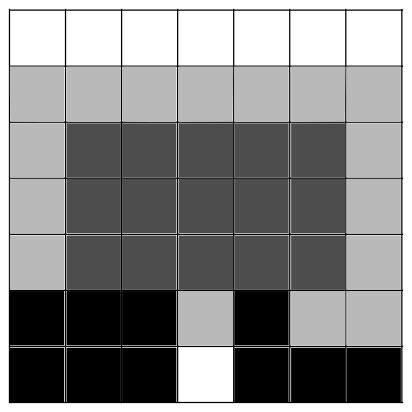} \\ (b)
        \end{minipage}
        \begin{minipage}[t]{0.35\columnwidth}
            \centering
            \includegraphics[width = \textwidth]{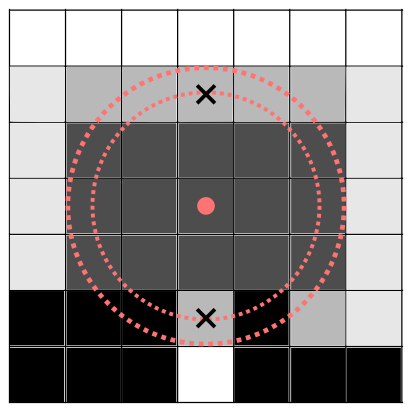} \\ (c)
        \end{minipage}
        \hfill
        \begin{minipage}[t]{0.35\columnwidth}
            \centering
            \includegraphics[width = \textwidth]{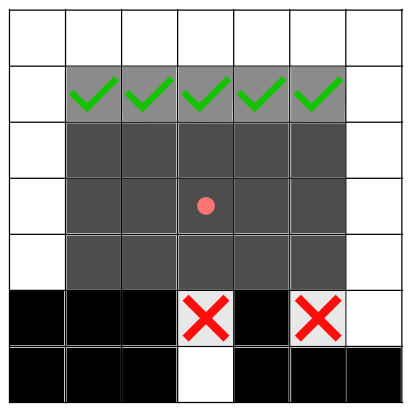} \\ (d)
        \end{minipage}
    \end{center}
    \caption{Outline of one step of the cluster growing algorithm. (a) Cluster shape of the current iteration. (b) Find direct neighbors of the cluster. (c) Choose all neighbors which make the cluster compact. (d) Only keep the voxels that preserve the convexity.}
    \label{fig:cluster_growing}
\end{figure}

The next step in the \emph{Topomap} pipeline is to grow a set of compact, convex clusters based on the binarized TSDF reconstruction of the environment.
Enforcing a compact growing leads to a sphere-like expansion of the clusters (i.e. a similar spread in all directions) instead of making it dependent on the cubic shape of the voxels or the orientation of the voxel grid.

The cluster growing algorithm initializes a first cluster center voxel at a random position along the explorer trajectory, because there we have the highest certainty about free space.
Then we start an iterative growing of the initialized cluster, which consists of three main steps (\reffig{fig:cluster_growing}):
\begin{enumerate}
\item Find all directly adjacent non-occupied voxels for the current voxel cluster.
\item Perform Principal Component Analysis and find the principal axes of the current cluster shape that contain most of the currently included voxels (we use the threshold of $98\%$), assuming an ellipsoidal shape for the clusters.
If the smallest half axis of this ellipse has a length of $r_\mathrm{min}$, we only allow voxels up to a distance of $r_\mathrm{min} + \delta$ to the cluster center to be added to the cluster (compact candidates).
\item  Only add these compact candidates which ensure that the convex hull of the current cluster does not include any obstacle voxels.
This is done by checking if the rays from all candidate voxels to the cluster voxels are obstacle-free.
\end{enumerate}

The growing of a single cluster terminates after no more voxels have been added.
This means that there are either no more compact candidates found or that the existing cluster is no longer convex when adding these candidates.
After this, we start growing the other clusters in the same way until the whole explorer trajectory is contained within the clustered space.

\subsection{Convex Cluster Merging}
\label{sec:cluster_merging}

The algorithm described in the previous section yields a relatively high number of compact, convex regions which are conservative in the sense that they do not include any obstacle voxels and have a similar expansion in all directions.
We now introduce a second step, where regions are merged if the convex hull of the combined region contains a low number of obstacle voxels.
Our algorithm is inspired by \emph{dynamic region merging}~\cite{Peng2011}, which first over-segments an image into small and conservative regions (superpixels), and then merges similar segments iteratively.
This two-step procedure makes the final segmentation less prone to noisy input data, as the merging step is based on a good but over-conservative pre-segmentation of the input pixels/voxels.

The proposed algorithm is iterative and repeatedly attempts to merge neighboring candidate pairs of clusters.
In each iteration, it starts by searching for all merge candidates, that is pairs of clusters which are directly adjacent.
Then, it iterates through all tentative cluster merge pairs in a randomized fashion.
For each cluster pair, it computes the combined convex hull of this pair using the Quickhull algorithm~\cite{Barber1996}. 
The cluster pair is merged if the relative number of contained occupied voxels (i.e. obstacles) within this convex hull is smaller than some set \emph{obstacle ratio threshold} (we set it to $1$-$5\%$ in our experiments).
Increasing this value will lead to a lower number of clusters, but will indeed leave more responsibility to a local planner.
We elaborate on the influence of this parameter in \refsec{sec:topomap_creation_evaluation}.
The procedure is terminated when no more merges were performed based on the merging criterion.

\subsection{Topological Navigation}
\label{sec:topo_navigation}

After building a convex cluster representation of the environment, we can proceed with the topological navigation.
We interpret the convex clusters as the vertices of our topological graph.
Similarly, the topological edges are created whenever two clusters are adjacent (portals).
\reffig{fig:example_idea:navi_graph} outlines our approach of path planning on a topological map.

Let us consider a simple A-to-B path planning case:
both A and B have to be located within the clusters as we have no information about the space outside of them.
We start by building a navigation graph by connecting the portal centers of all topological vertices.
Additionally we connect A and B to the portal centers of their corresponding topological vertices.
In order to get the shortest path from A to B based on this graph, we can perform an A* search.
The path planning algorithm would then plan a path where the agent moves from A to the portal, then starts traversing intermediate clusters until it reaches the cluster that contains B, where it can move directly from the portal to the desired destination.

\section{Experimental Evaluation}
\label{sec:experimental}

In this section, the evaluation of three main parts of \emph{Topomap} and the results of the entire framework are presented.
\refsec{sec:tsdf_map_evaluation} evaluates the performance of TSDF integration based on sparse visual 3D landmarks and compares it to the approach based on dense reconstruction using a stereo camera.
It indicates the potential compromises we are making by avoiding the use of any expensive hardware or significant computational power.
Then, we demonstrate the performance of the cluster creation algorithm on multiple real world environments in \refsec{sec:topomap_creation_evaluation}.
Finally, \refsec{sec:path_planning_evaluation} compares our topological planner to RRT*, a state-of-the-art planner that is commonly used within the robotics community.

We acquired our datasets using a synchronized visual-inertial sensor~\cite{Nikolic2014} and the experiments were performed using a dual-core Intel i7-7600U ($2.8$ GHz) processor.
The SLAM maps are built by first running a visual-inertial odometry pipeline based on~\cite{Leutenegger2015} and then running global bundle adjustment and loop closure using \emph{maplab}, an open source mapping framework~\cite{schneider2018maplab}.
%

\subsection{TSDF Maps from Visual Landmarks}
\label{sec:tsdf_map_evaluation}

In the previous sections we proposed ray tracing free space from sparse SLAM landmarks that significantly reduces computational requirements and simplifies the sensor setup, but it might affect the quality of the TSDF reconstruction.
Capturing both the free space and occupied areas correctly is essential for a reliable creation of the convex clusters and navigation in the subsequent steps of our algorithm.
We would therefore like to evaluate the TSDF maps of the proposed approach and compare them with TSDF maps that were created from dense stereo images.
By employing the semi-global matching algorithm~\cite{Hirschmuller2008} implemented in OpenCV~\cite{Itseez2015}, we get dense depth images from the $10\,\mathrm{cm}$ baseline stereo camera.
These are integrated into a TSDF from their corresponding observer poses in the same way as the 3D landmarks (see \refsec{sec:occupancy}) using the identical integration parameters.
The most important parameters are the maximum ray length and the truncation distance, which we set to $4.0$-$7.0\,\mathrm{m}$ and $0.1$-$0.5\,\mathrm{m}$, respectively.

\begin{figure}[tpb]
    \begin{center}
        \begin{minipage}[t]{0.48\columnwidth}
            \centering
            \includegraphics[width = \textwidth]{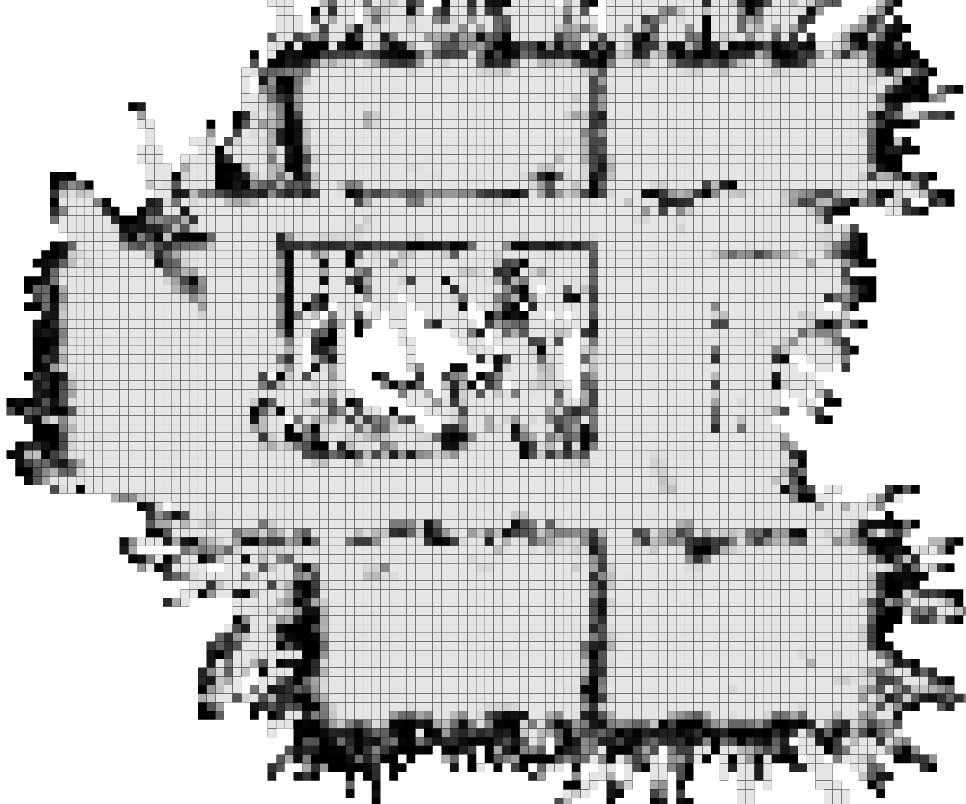} \\ (a) Depth map integration.
        \end{minipage}
        \hfill
        \begin{minipage}[t]{0.48\columnwidth}
            \centering
            \includegraphics[width = \textwidth]{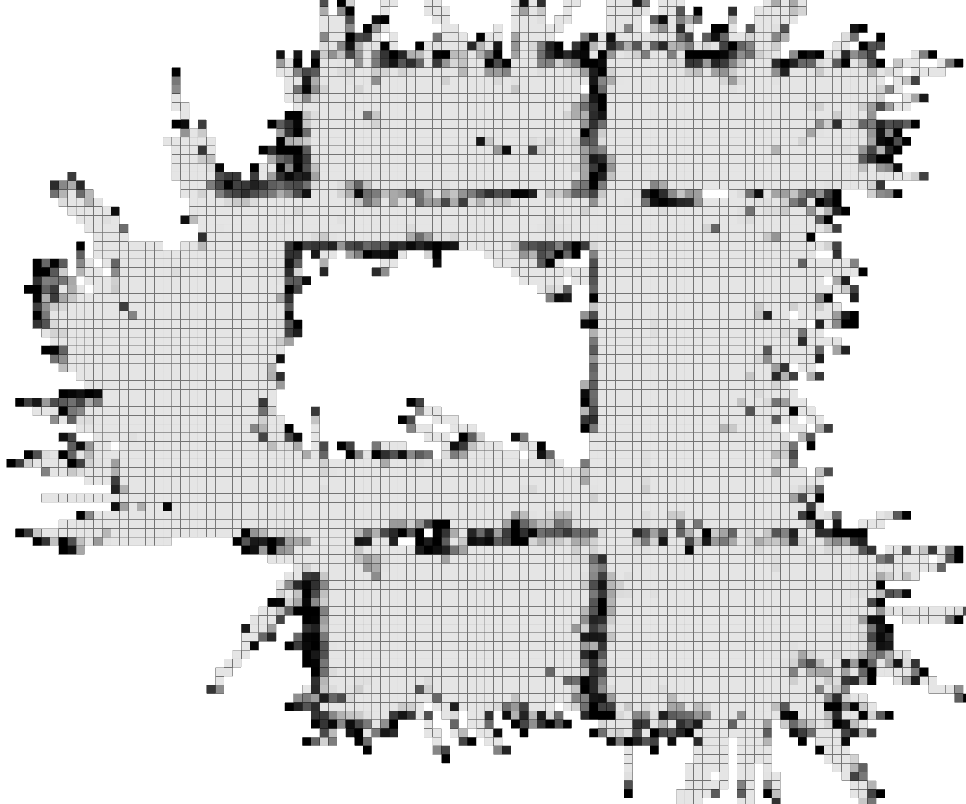} \\ (b) Sparse landmark integration.
        \end{minipage}
    \end{center}
    \caption{Slice of a 3D TSDF reconstruction of an office environment.
    The voxel size is $0.25\,\mathrm{m}$.
    Integrating the $1322$ stereo images takes $43.3\,\mathrm{s}$ and produces a TSDF map of $65'231$ voxels, whereas the integration of the $177'269$ landmarks lasts for $2.1\,\mathrm{s}$ and the corresponding map contains $39'791$ voxels.
    Observe how using sparse landmarks for TSDF construction preserves most of the relevant environment structure.}
    \label{fig:tsdf_dense_sparse}
\end{figure}

A qualitative comparison between sparse visual landmarks and depth maps to create a TSDF map is presented in \reffig{fig:tsdf_dense_sparse}.
Even if the TSDF map based on the landmarks is sparser and partially occupied by outlier voxels in some of the free space areas, it still contains the relevant structure which can be inferred from the dense map (walls, corridors, doorways).
In fact, the \emph{Topomap} framework handles outliers and missing information in the subsequent stages by filtering out small connected components in the occupancy map and by enforcing convexity for all added voxels during the growing process.
Convexity prevents clusters from growing through or around obstacles (e.g. walls) even if some occupancy information is missing.

\begin{figure}[tpb]
    \centering
    \includegraphics[width=\columnwidth]{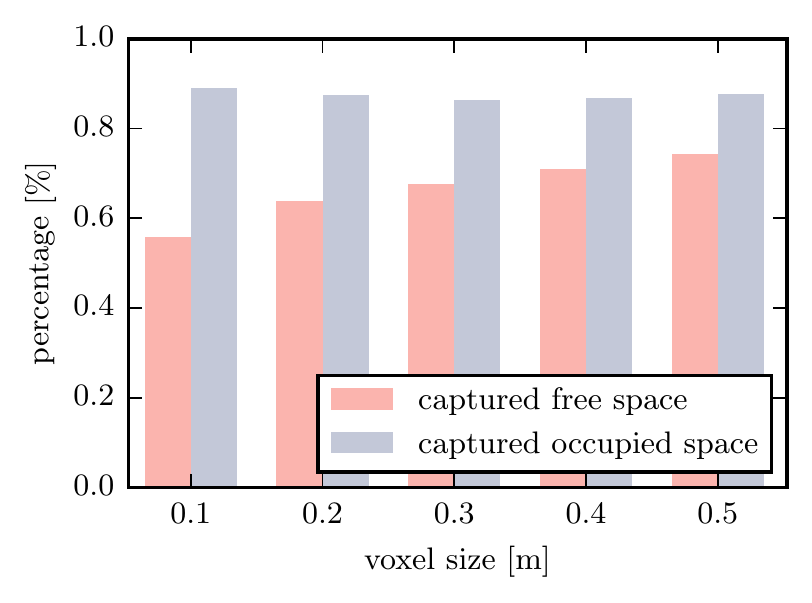}
    \caption{Influence of the voxel size on the amount of captured free/occupied space in the sparse TSDF map.
    The dense TSDF map is taken as a reference: For each free voxel in the dense map, we check if the corresponding voxel in the sparse map is free as well.
    In order to get the captured occupied space, we also take into account unmapped space in the sparse map.
    This is reasonable as during the voxel growing, unmapped space is treated as occupied space.
    }
    \label{fig:tsdf_map_evaluation}
\end{figure}

In the next step, we want to evaluate quantitatively how much free space can be captured using the sparse visual landmarks compared to the dense maps generated from a stereo rig.
Here, we take the dense TSDF map as a reference.
Obviously, this is only an approximation, but should give a good insight about how well the landmark integration performs in real world scenarios.
\reffig{fig:tsdf_map_evaluation} shows the percentage of captured free and occupied space for five voxel resolutions.
More free space can be captured if we increase the voxel size, but at the same time, we will also get larger discretization errors (e.g. small obstacles will not be captured in the occupancy map).
Evidently, this latter effect is not captured in the shown plot, as the dense TSDF map suffers from the same effects.

\subsection{Topological Map Creation}
\label{sec:topomap_creation_evaluation}

\begin{figure*}[tpb]
    \centering
    \includegraphics[width=\textwidth]{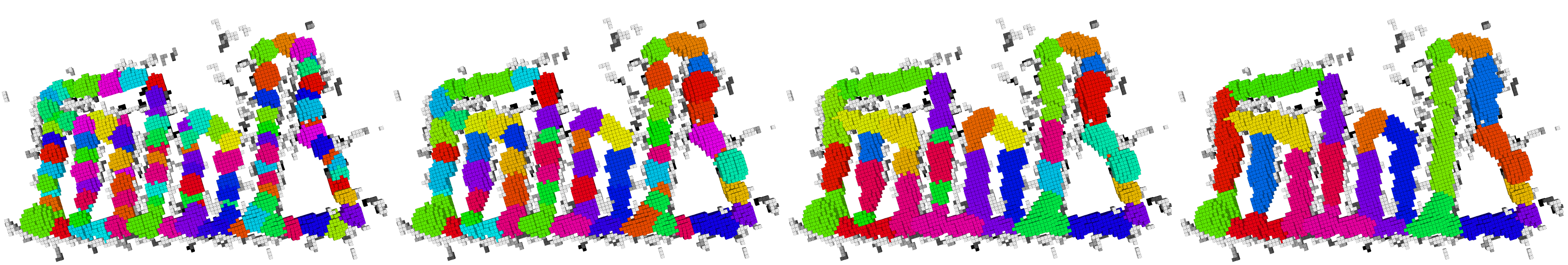}
    \caption{Iterative cluster merging demonstrated in a \emph{warehouse} setting.
    Starting with a large number of small and compact clusters, we merge cluster pairs which contain only a low number of obstacle voxels within their combined convex hull.
    The initial $105$ clusters are reduced to $24$ clusters within $3$ merging steps, and the number of topological edges is reduced from $121$ to $29$.
    In this example, we set the obstacle ratio threshold to $5\%$.
    }
    \label{fig:cluster_merging}
\end{figure*}

In this section, we want to evaluate the core part of \emph{Topomap} -- growing and merging of the voxel clusters.
\reffig{fig:cluster_merging} shows the cluster arrangement after the growing step and the subsequent iterations of the merging algorithm in a warehouse environment.
While the cluster growing step expands the clusters within the free space, the merging procedure leads to a substantial complexity reduction of the topological map structure.
It is worth to emphasize how the algorithm captures the complex topology of this environment by combining multiple small compact clusters into larger ones along corridors, but preserves a more fine-grained clustering in the corners.

\begin{figure}
    \centering
    \begin{subfigure}[b]{0.9\columnwidth}
        \centering
        \includegraphics[width=1.0\columnwidth]{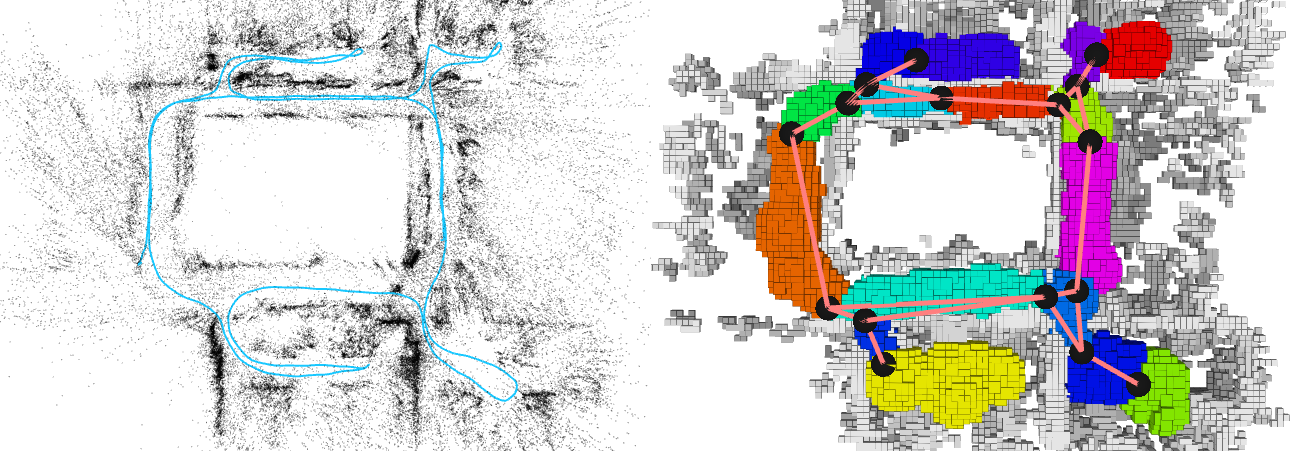}
        \caption{Office dataset, $20\,\mathrm{m}$ trajectory}
        \label{fig:benchmark_datasets:office}
    \end{subfigure}
    
    \begin{subfigure}[b]{0.9\columnwidth}
        \centering
        \includegraphics[width=1.0\columnwidth]{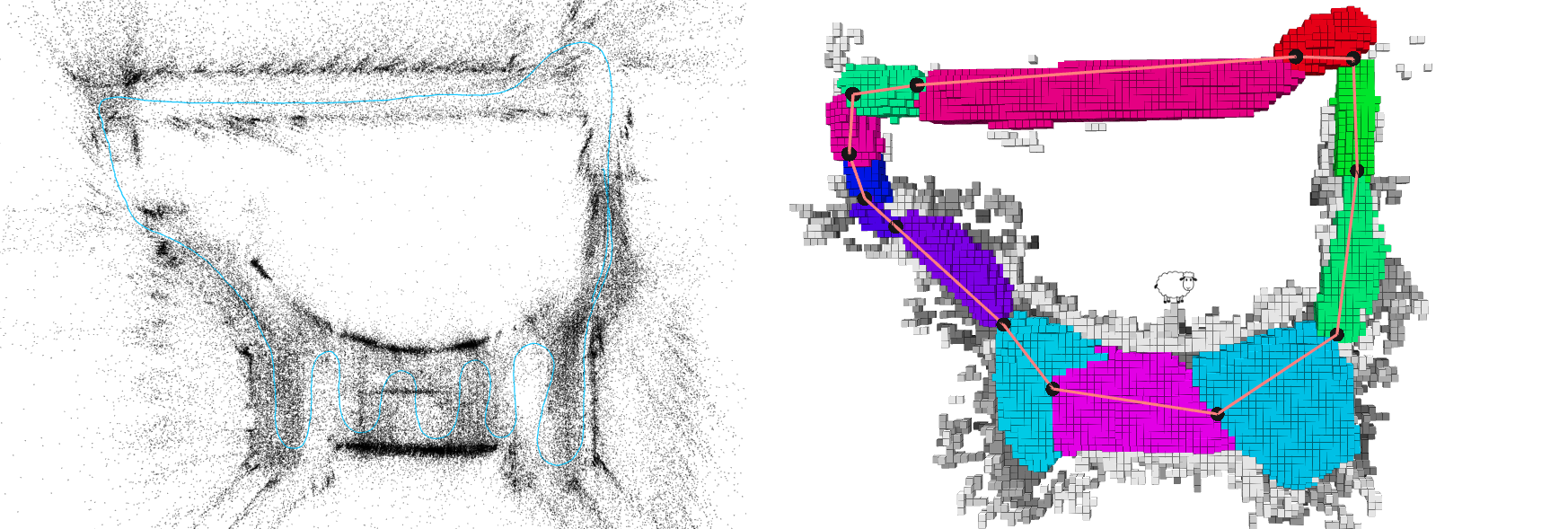}
        \caption{Open space, $50\,\mathrm{m}$ trajectory}
        \label{fig:benchmark_datasets:open}
    \end{subfigure}
    
    \begin{subfigure}[b]{0.9\columnwidth}
        \centering
        \includegraphics[width=1.0\columnwidth]{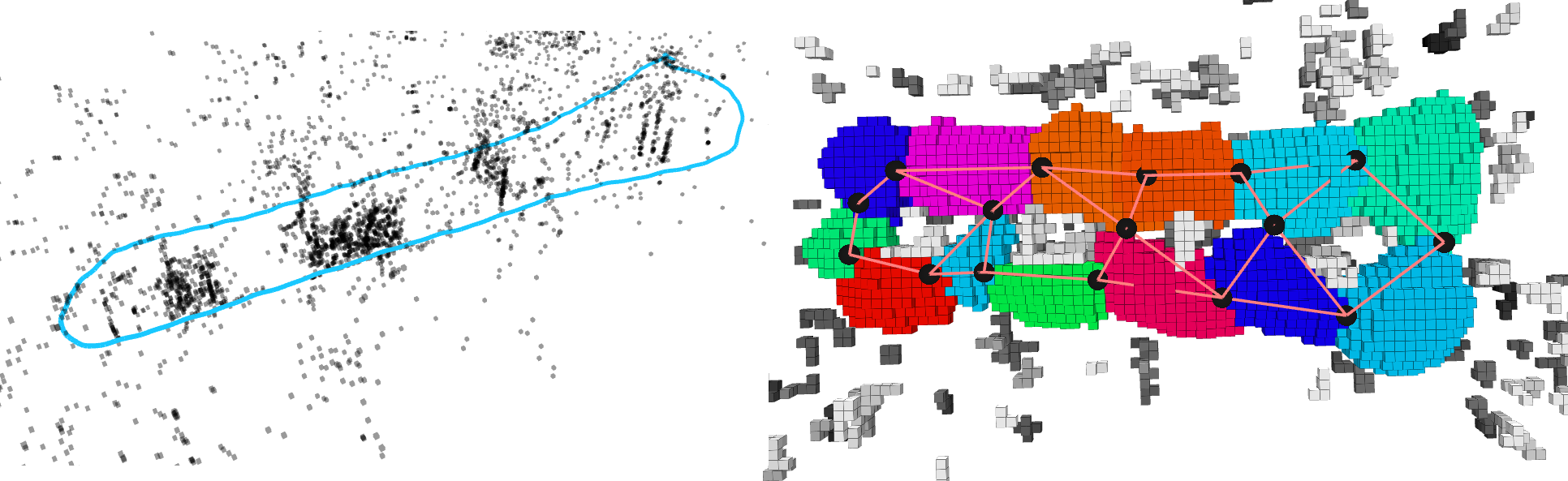}
        \caption{Pillars dataset, $10\,\mathrm{m}$ trajectory }
        \label{fig:benchmark_datasets:pillars}
    \end{subfigure}
    \caption{
    SLAM maps (left) and their corresponding topological maps (right).
    (a)~Office: Typical topological mapping dataset consisting of narrow corridors and clearly separated rooms.
    (b)~Open space: The large open spaces of this dataset are correctly represented by convex clusters.
    (c)~Pillars: The ray traced empty space between the clusters is sufficient to introduce topological edges and enable path planning in these areas where the explorer never traversed.
    }
    \label{fig:benchmark_datasets}
\end{figure}

To give more insights into the results of the clustering algorithm, we also showcase the output from \emph{Topomap} for three different datasets in \reffig{fig:benchmark_datasets}:
The \textit{office} dataset constitutes a typical example of a structured environment, which is segmented into clusters in a similar fashion to what a human would typically do (separation into \emph{rooms} and \emph{corridors}).
The second example, \textit{open space}, is more demanding as it consists of large open spaces at the one hand and of some more narrow passages on the other (top part of the map).
Note how our topological map representation succeeds to simplify the map given the winding explorer trajectory.
Clearly, the merging algorithm did well by creating rather large clusters in the open space part of the dataset.
Finally, the \textit{pillars} dataset highlights how \emph{Topomap} can infer traversable areas that cannot be deduced purely from the explorer trajectory.
This means that a robot that uses this topological map could use the passages in between the pillars even if they have never been traversed by the explorer.
This would not be possible for a teach-and-repeat navigation strategy.

\begin{figure}[tpb]
    \centering
    \includegraphics[width=\columnwidth]{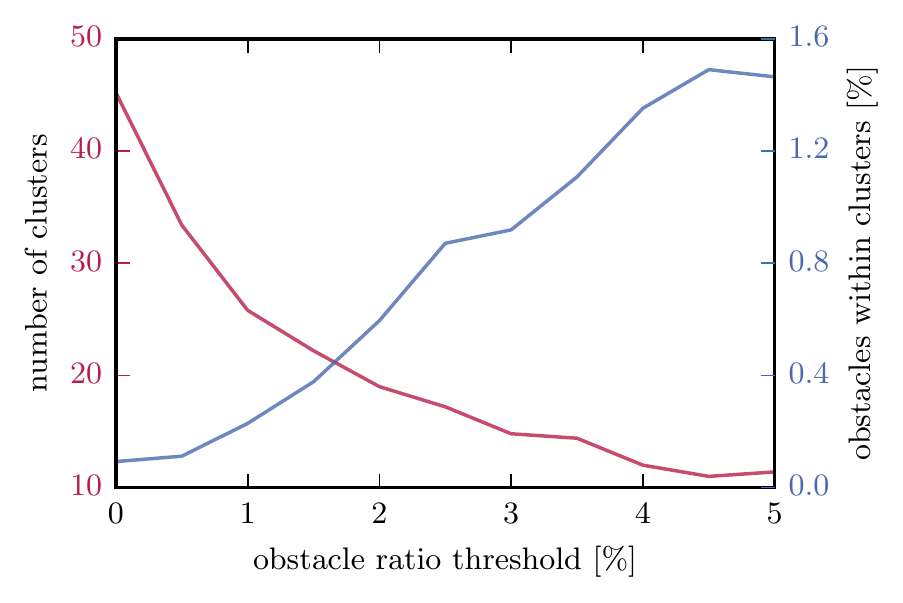}
    \caption{An evaluation of the obstacle ratio threshold parameter for the \emph{open space} dataset.
    As expected, a higher value of this parameter leads to a lower number of final clusters (red curve) as many of them are merged.
    On the other hand, the blue curve expresses how many obstacle voxels are contained within the voxel clusters.
    It increases if we leverage the condition of contained obstacle voxels.
    }
    \label{fig:topomap_evaluation}
\end{figure}

The \emph{Topomap} algorithm does not require a large set of environment specific parameters or tedious fine-tuning for a specific use case. 
We would like, however, to highlight a parameter that affects both the robustness towards outliers and the accuracy of the topological map.
\reffig{fig:topomap_evaluation} demonstrates the influence of the \emph{obstacle ratio threshold} (introduced in \refsec{sec:cluster_merging}) on the segmentation result.
This parameter gives us control over the following trade-off:
Larger values of the \emph{obstacle ratio threshold} will reduce the complexity of the topological map by accepting small obstacles to be present within the merged clusters.
These violations of the requirement that clusters are completely empty shifts part of the planning responsibility to the local planner and requires a local collision avoidance algorithm to circumvent these obstacles within a cluster.
Keeping the values of this parameter low will have an opposite effect -- only few obstacle voxels will be allowed within the clusters which will lead to a large number of small clusters.
It will therefore be harder to tune this parameter within large spaces where we encounter varying obstacle/free space configurations.

\begin{table}[]
\centering
\caption{Timing and storage requirement statistics of the \emph{Topomap} pipeline for the datasets evaluated in this paper. The volume within the brackets shows the volume of all mapped voxels within the TSDF map.
All voxel sizes were set to $0.25\,\mathrm{m}$.
The storage requirements of our approach are significantly reduced when compared to the full TSDF.
}
\label{tab:topomap_statistics}
\resizebox{\columnwidth}{!}{%
\begin{tabular}{lcccccc}\toprule
                                & \multicolumn{2}{c}{\textbf{computation time {[}s{]}}} &  & \multicolumn{3}{c}{\textbf{storage requirements {[}kB{]}}} \\
                                & TSDF                    & clusters                    &  & full clusters         & convex hulls        & TSDF        \\ \cline{2-3} \cline{5-7} 
multifloor ($2209\,\mathrm{m}^3$) & 4.1                    & 86.1                        &  & 567                   & 249                  & 4320          \\
warehouse ($895\,\mathrm{m}^3$)  & 0.4                    & 11.2                          &  & 147                    & 110                  & 1784          \\
office ($622\,\mathrm{m}^3$)     &  2.1                   & 9.9                          &  & 123                    & 73                  & 1162          \\
open space ($1464\,\mathrm{m}^3$) & 2.7                    & 42                          &  & 379                    & 150                  & 2932          \\
pillars ($269\,\mathrm{m}^3$)    & 0.46                     & 3.1                          &  & 51.9                    & 62.7                  & 607         
\\\bottomrule
\end{tabular}%
}
\end{table}

We claim that \emph{Topomap} is particularly useful in the applications that put constraints on the computational power or limit the available perception hardware.
Special care was taken to guarantee that the framework uses a limited amount of resources, including the storage requirements of the topological map.
\reftab{tab:topomap_statistics} summarizes some statistics about the topological map creation for the datasets presented throughout this paper.
Having the SLAM map as an input to our system, we could create topological maps for most of these datasets within less than a minute.
The storage requirements for the topological map are low as only the convex hulls of the voxel clusters are stored.
Obviously, this compact representation has also the potential to over-simplify the environment and we may lose important details in some cases.

\subsection{Path Planning}
\label{sec:path_planning_evaluation}

\begin{figure}[tpb]
    \centering
    \includegraphics[width=\columnwidth]{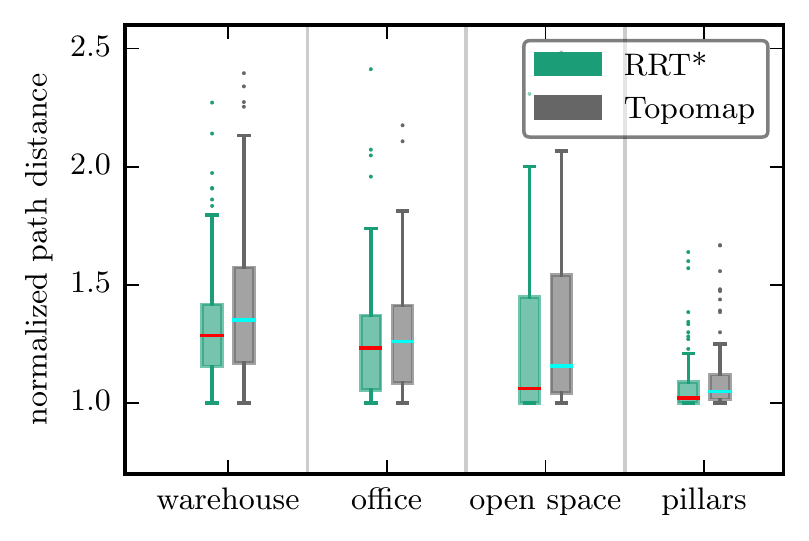}
    \caption{A comparison of the normalized path distances of RRT* operating on a TSDF map from dense stereo images and our proposed topological planner ($100$ runs per dataset). 
    The path distances are normalized by the direct line distance.
    Our lightweight topological planner generates paths just marginally longer than RRT*, but requires much simpler computations (A* on a small navigation graph).}
    \label{fig:planner_comparison}
\end{figure}

The proposed topological mapping concept is primarily targeting navigation and path planning.
Below, we present an evaluation of the entire pipeline that includes both the assessment of the generated paths as well as a deployment of the system on an real robotic platform.

First, we compare \emph{Topomap} to the RRT* planner from OMPL~\cite{Sucan2012}, which is provided a TSDF map from stereo images.
We sample 100 trajectories with random start and goal positions for both planners and compare the path distances normalized by the direct line distance (see~\reffig{fig:planner_comparison}).
The planning time of the RRT* planner is set to $2\,\mathrm{s}$, which led to successful paths given the complexity of our environments and the voxel resolution.
In general, the path lengths generated by the topological planner are slightly longer, but at the same time, our A* planning time is drastically lower than for RRT* (typically in the order of some $100\,\mathrm{\mu s}$).
%
This directly corresponds to our goal to replace demanding algorithms with a lightweight counterpart with only a marginal quality loss.

\begin{figure}[tpb]
    \centering
    \includegraphics[width=\columnwidth]{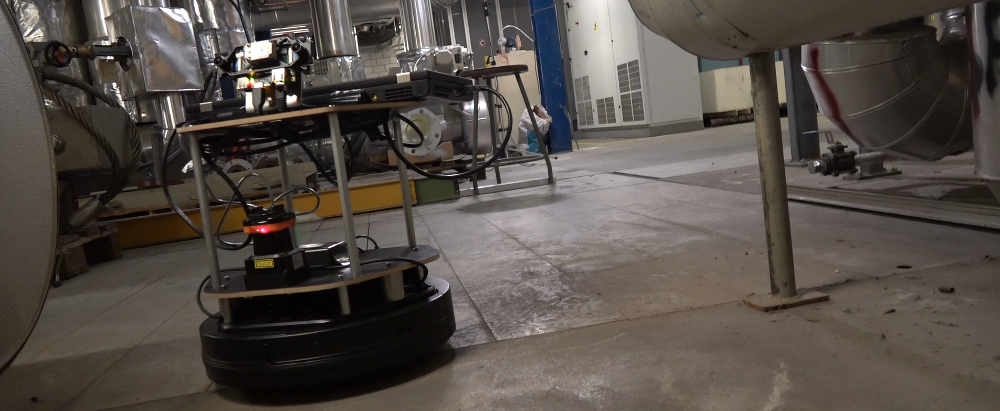}
    \caption{Setup of the Turtlebot experiments. 
    The VI sensor~\cite{Nikolic2014} is used for global localization within the SLAM map, and a laser is used for local obstacle avoidance only using the dynamic window approach~\cite{Fox1997}.}
    \label{fig:turtlebot_setup}
\end{figure}

Secondly, we have integrated our proposed navigation system on a Turtlebot robot equipped with a VI sensor, and successfully performed path planning tasks within a semi-structured industrial site.
In a first step, the robot performed global localization within the visual SLAM map using the approach described in \cite{Lynen2015}, and was then given a target position.
The topological planner then computed the fastest path to the given location, and Turtlebot successfully completed a trajectory of approximately $15\,\mathrm{m}$, using a 2D laser for local obstacle avoidance only.
These experiments proved the usefulness of our system on a mobile platform equipped only with a camera and a computationally constrained processing unit.
A video footage demonstrating the Turtlebot experiment is provided as a supplementary material to this paper\footnote{\url{https://youtu.be/UokjxSLTcd0}}.

\section{Conclusions}

In this paper, we have presented \emph{Topomap}, a novel framework for creating versatile topological maps and reliable navigation therein.
Our approach can handle noisy and sparse visual measurements, which significantly reduces the hardware requirements when compared with state-of-the-art approaches.
The core component of the proposed system is a voxel based growing and merging algorithm, which segments the free space into convex clusters.
This enables path planning algorithms that are orders of magnitude faster than conventional grid based planners.
Additionally, the chosen structure of the topological map makes the resulting maps very compact.

The evaluations of \emph{Topomap} demonstrate that it is possible to reliably build TSDF maps from sparse vision-based measurements.
We also have proved that the results of the framework do not exhibit any significant quality loss when compared to RRT* operating on a dense TSDF map.
Finally, the system was successfully deployed on a mobile robotic platform.
We believe the results of this work will be interesting for everyone working on the navigation of mobile platforms within large-scale environments, and where the computational resources, size or weight are limited.

For future work, we plan to evaluate our approach on some larger real world scenarios and we would like to integrate it on a flying platform, which would exploit the full 3D capabilities of this framework.
Furthermore, we intend to include semantic information in the topological maps, as this has the potential of pushing boundaries of robotic autonomy even further.


\section*{Acknowledgement}

We would like to thank Helen Oleynikova for fruitful discussions about navigation, path planning and real challenges of mobile robotics.§
The research leading to these results has received funding from Google Tango.

\balance

\bibliographystyle{IEEEtran} 
\bibliography{references}

\end{document}